\documentclass[sigconf,natbib=true]{acmart}
\AtBeginDocument{%
  }

\copyrightyear{2026}
\acmYear{2026}
\setcopyright{None}\acmConference{}{}{}
\acmBooktitle{}
\acmDOI{}
\acmISBN{}

\usepackage{xcolor}
\usepackage{multirow}
\usepackage{tabularx}
\usepackage{adjustbox}
\usepackage[normalem]{ulem}
\useunder{\uline}{\ul}{}
\usepackage{makecell}
\usepackage{color}
\usepackage{subfig}
\usepackage{graphicx}
\usepackage{colortbl}
\usepackage{enumitem}

\usepackage{subcaption}   
\usepackage{float} 

\definecolor{lightgray}{rgb}{0.95, 0.95, 0.95}
\definecolor{lightblue}{rgb}{0.85, 0.92, 1.0}
\newcommand{\deltacell}[2]{$#1_{#2\%\downarrow}$}
\newcommand{\maxdeltacell}[2]{$#1_{#2\%\downarrow}$}
\newcommand{\seconddeltacell}[2]{$#1_{#2\%\downarrow}$}

\begin{document}

\title{Exploring Knowledge Conflicts for Faithful LLM Reasoning: Benchmark and Method }

\author{Tianzhe Zhao}
\affiliation{%
  \institution{School of Computer Science and Technology, Xi'an Jiaotong University}
  \city{Xi'an}
  \country{China}
}
\email{ztz8758@foxmail.com}

\author{Jiaoyan Chen}
\affiliation{%
  \institution{Department of Computer Science, The University of Manchester}
  \city{Manchester}
  \country{United Kindom}
}
\email{jiaoyan.chen@manchester.ac.uk}

\author{Shuxiu Zhang}
\affiliation{%
  \institution{Hunan University}
  \city{Changsha}
  \country{China}
}
\email{zsx@hun.edu.cn}

\author{Haiping Zhu}
\affiliation{%
  \institution{School of Computer Science and Technology, Xi'an Jiaotong University}
  \city{Xi'an}
  \country{China}
}
\email{zhuhaiping@xjtu.edu.cn}

\author{Qika Lin}
\authornote{Corresponding authors.}
\affiliation{%
  \institution{National University of Singapore}
  \city{Singapore}
  \country{Singapore}
}
\email{linqika@nus.edu.sg}

\author{Jun Liu}
\authornotemark[1]
\affiliation{%
  \institution{School of Computer Science and Technology, Xi'an Jiaotong University}
  \city{Xi'an}
  \country{China}
}
\email{liukeen@xjtu.edu.cn}

\renewcommand{\shortauthors}{Tianzhe Zhao et al.}


\begin{abstract}
Large language models (LLMs) have achieved remarkable success across a wide range of applications especially when augmented by external knowledge through retrieval-augmented generation (RAG).
Despite their widespread adoption, recent studies have shown that LLMs often struggle to perform faithful reasoning when conflicting knowledge is retrieved.
However, existing work primarily focuses on conflicts between external knowledge and the parametric knowledge of LLMs, leaving conflicts across external knowledge largely unexplored.
Meanwhile, modern RAG systems increasingly emphasize the integration of unstructured text and (semi-)structured data like knowledge graphs (KGs) to improve knowledge completeness and reasoning faithfulness.
To address this gap, we introduce \textbf{ConflictQA}, a novel benchmark that systematically instantiates conflicts between textual evidence and KG evidence.
Extensive evaluations across representative LLMs reveal that, facing such cross-source conflicts, LLMs often fail to identify reliable evidence for correct reasoning.
Instead, LLMs become more sensitive to prompting choices and tend to rely exclusively on either KG or textual evidence, resulting in incorrect responses.
Based on these findings, we further propose \textbf{XoT}, a two-stage explanation-based thinking framework tailored for reasoning over heterogeneous conflicting evidence, and verify its effectiveness with extensive experiments.
\end{abstract}




\maketitle

\section{Introduction}
Large Language Models (LLMs) have demonstrated remarkable achievements in a wide range of applications~\cite{naveed2025comprehensive, zhao2023survey,DBLP:journals/tkde/XuLHZLC25}. 
Retrieval-Augmented Generation (RAG) further extends LLMs by integrating external knowledge sources, enabling responses to be grounded in retrieved content rather than generated solely from the model’s internal parameters~\cite{gao2023retrieval, lewis2020retrieval, fan2024survey}. 
By leveraging multiple heterogeneous knowledge sources, such as unstructured textual documents alongside (semi-)structured knowledge graphs (KGs)~\cite{lee2025hybgrag} and tables~\cite{dong2025reasoning}, LLMs are better equipped to mitigate hallucination and support knowledge-intensive tasks.

Despite the growing adoption of RAG systems, their reliance on external knowledge also introduces new challenges.
In real-world scenarios, external knowledge may be outdated, noisy, or intentionally manipulated, leading to conflicting or contradictory evidence during LLM reasoning~\cite{xu-etal-2024-knowledge-conflicts, su2024textttconflictbank, liang2025saferag}. 
To deal with such issues, some recent studies have begun to investigate the faithfulness of LLM reasoning in the presence of conflicting knowledge~\cite{ICLR2025_48404cd9, wang2025astute,jin-etal-2024-tug}.
However, these existing studies predominantly focus on inconsistencies between external evidence and LLMs’ parametric knowledge~\cite{jin-etal-2024-tug, su2024textttconflictbank}, or consider conflicts among external evidence that are synthetically constructed within a single knowledge source~\cite{ICLR2025_48404cd9}.
In contrast, cross-source conflicts from heterogeneous external knowledge sources remain largely unexplored.

\begin{figure}[t]
  \centering
  \includegraphics[width=0.98\linewidth]{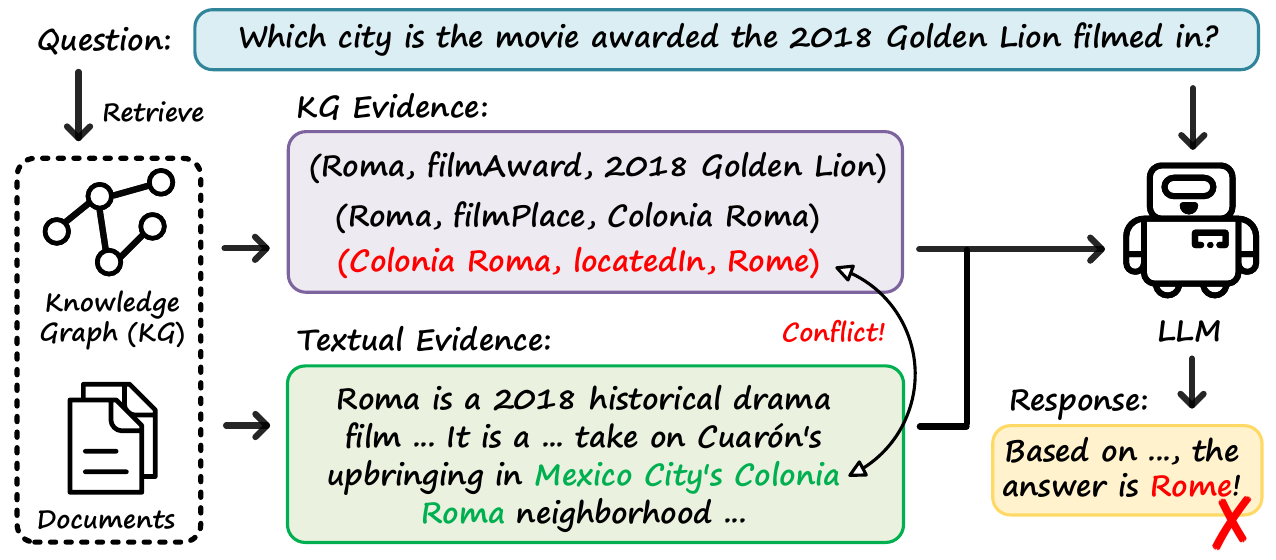}
  \caption{Demonstration of  cross-source knowledge conflict in LLM reasoning with RAG.}
  \Description{}
  \label{fig:intro}
  \vspace{-0.4cm}
\end{figure}

To address this gap, in this study we investigate conflicts that arise between heterogeneous external knowledge sources, using textual documents and KGs, two of the most widely adopted external sources in modern RAG systems, as a case.
Textual documents (e.g., Wikipedia articles and passages) are most commonly retrieved to provide informative, context-rich descriptions,
while KGs, which represent relational facts in the form of triples, i.e., \textit{(head, relation, tail)}, have also been increasingly incorporated into RAG systems to support complex reasoning, especially in domains that require high trustworthiness, such as legal judgment and medical diagnosis~\cite{peng2025graph, procko2024graph, barron2025bridging, zhao2025medrag}.
Beyond serving as standalone knowledge sources, recent studies have also highlighted the necessity of jointly integrating textual documents and KGs to provide more comprehensive knowledge for LLM reasoning in real-world scenarios~\cite{ma2025thinkongraph, wu2024stark, lee2025hybgrag, li2024chainofknowledge, xia2025knowledge}.
Yet, how LLMs assess and resolve inconsistencies that arise between textual descriptions and KG triples under such settings remains unclear.
As shown in Figure~\ref{fig:intro}, when answering the question 
\textit{``Which city is the movie awarded the 2018 Golden Lion filmed in?''}, 
the KG fact about the location of \textit{Colonia Roma} conflicts with textual evidence (e.g., \textit{"Mexico City"}).
As a result, LLMs may struggle to identify the more reliable evidence and generate satisfactory responses.
However, there is a shortage of benchmarks specifically designed to study such conflicts in multiple heterogeneous sources, which hinders the systematic evaluation of LLMs’ reasoning behavior and the development of more faithful reasoning methods.

In this work, we construct a novel question answering (QA) benchmark, called ConflictQA, to systematically investigate faithful LLM reasoning under conflicting evidence retrieved from heterogeneous sources.
Following previous works~\cite{su2024textttconflictbank, ICLR2025_48404cd9}, ConflictQA operates in a post-retrieval manner, where each question is paired with evidence from both textual documents and KGs, provided respectively as passages and sets of triples. 
To simulate conflicts, we instantiate textual and KG evidence that provides inconsistent cues for the same fact that the answer depends on.
As exemplified in Figure~\ref{fig:intro}, the evidence in triples and texts lead to different answers.
Motivated by real-world reasoning requirements, ConflictQA covers two settings: (i) Non-complementary (Non-COMP) cases, where evidence from either textual documents or KGs alone is adequate to answer the question, 
and (ii) complementary (COMP) cases, where answering the question requires jointly reasoning over both textual and KG evidence, as illustrated in Figure~\ref{fig:intro}, where the award information is only available in the KG evidence.
Within each setting, we further distinguish which source provides the negative evidence supporting incorrect answers, enabling fine-grained analysis of how LLMs assess evidence reliability and resolve conflicts.

With the ConflictQA benchmark, we conduct comprehensive evaluations to examine LLM reasoning behavior under conflicts arising from heterogeneous external sources.
We evaluate 12 representative models, covering both general-purpose and reasoning-specialized LLMs.
Our results reveal a consistent limitation across models: 
when confronted with conflicting evidence from different sources, LLMs fail to reliably determine which evidence to trust, leading to incorrect answers in both non-complementary and complementary settings.
Through analyses on evidence ordering and prompting strategies, we also observe an interesting tendency:
LLMs exhibit a systematic bias toward believing concise KG triples when prompted to directly generate answers, even when the triple is incorrect.
In contrast, chain-of-thought (CoT) prompting that elicits step-by-step reasoning may shift models’ decisions toward textual evidence.
These findings further suggest that current LLMs lack a robust mechanism for resolving cross-source conflicts or assessing evidence reliability.

Based on these observations, we further propose a two-stage explanation-based thinking framework (XoT) for more trustful LLM reasoning facing conflicting evidence.
Instead of prompting LLMs to produce answers directly,
XoT first encourages the model to enumerate all plausible answer candidates supported by different sources, together with explicit explanations for each candidate.
The final answer is then selected based on the aggregated explanations.
By separating candidate enumeration from answer selection, XoT helps LLMs mitigate premature bias toward a certain type of evidence and promotes more balanced reasoning when faced with conflicting information. 
Experiments on our ConflictQA benchmark demonstrates the effectiveness of XoT, as it achieves improved performance across a range of LLMs.
For example, in the complementary setting where KG evidence is misleading, XoT yields relative improvements of $20\%$ and $20\%$ on GPT-4o in terms of F1 score and exact match, respectively, compared to a conflict-aware QA prompt; On Open-Mistral-7B, the F1 score achieved by XoT is surprisingly almost threefold that obtained with the QA prompt.

In summary, the main contributions of this work lie in three aspects:

\noindent $\bullet$ We present the first systematic study of faithful LLM reasoning under cross-source knowledge conflicts, and construct ConflictQA, a novel benchmark that explicitly instantiates conflicts between textual and KG evidence\footnote{The ConflictQA benchmark is available at \url{https://github.com/Tianzhe26/ConflictQA}.}
.

\noindent $\bullet$ We conduct comprehensive evaluations of 12 representative LLMs on ConflictQA and analyze their reasoning behaviors under conflicting evidence, revealing consistent failure modes across different models and settings.

\noindent $\bullet$ Based on these analyses, we further propose XoT, a two-stage explanation-based thinking framework, that improves reasoning correctness under heterogeneous and misleading evidence across most evaluated models.

\section{Problem Definition}
\subsection{LLM Reasoning with Multiple Evidence}
We study a multi-source reasoning task for LLMs, where the model is provided with external evidence from heterogeneous knowledge sources, including textual documents and KGs.
For each question $q$, the available evidence consists of a set of textual passages $E_{\text{text}}$ and a set of KG triples $E_{\text{KG}}$.
The textual evidence is represented as a collection of passages, i.e.,
$E_{\text{text}} = \{p_1, p_2, \ldots, p_m\}$,
where each passage is a description relevant to the question.
The KG evidence is represented as a set of KG triples, i.e.,
$E_{\text{KG}} = \{(e^h_i, r_i, e^t_i)\}_{i=1}^n$,
where \(e^h_i\) and \( e^t_i\) denote the head and tail entities, respectively, and \(r_i\) denotes their relation.

Given $E_{\text{text}}$ and $E_{\text{KG}}$, the LLM is required to generate an answer set $\mathcal{A}$ grounded in the provided evidence. The process can be abstracted as:
\begin{equation}
    \mathcal{A} = f_{LLM}\big(\text{Prompt}(q, E_{\text{text}}, E_{\text{KG}})\big),
\end{equation}
where $f_{LLM}$ denotes an LLM, and $\text{Prompt}(\cdot)$ represents the prompting strategy that feeds the question and the evidence into the LLM.

\begin{figure*}[t]
  \centering
  \includegraphics[width=0.98\linewidth]{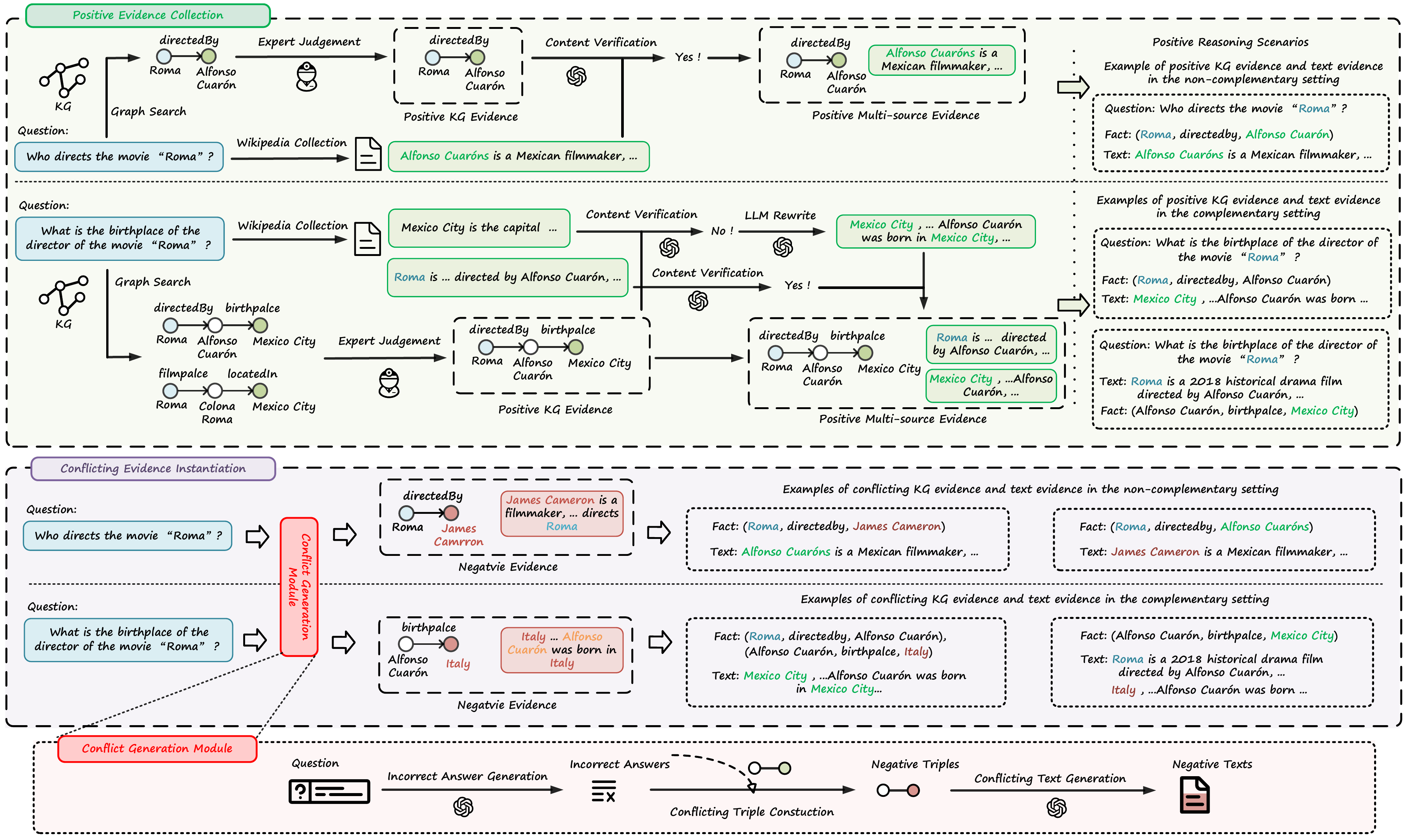}
  \caption{The pipeline for constructing the ConflictQA benchmark, mainly including positive and conflicting evidence construction as well as examples under both Non-COMP and COMP settings.}
  \Description{}
  \label{fig:bench_construct}
  \vspace{-0.2cm}
\end{figure*}

\subsection{LLM Reasoning with Conflicting Evidence}
\label{sec:confl}
In this work, we consider multiple evidence reasoning in which the provided evidence contains conflicting information and different evidence lead to different answers.
Specifically, we define two conflict scenarios based on which source provides misleading information.
One setting is \textit{TripleConf}, where all the textual evidence $E_{\text{text}}$ lead to the correct answer, while some KG evidence in $E_{\text{KG}}$ has inconsistent information and leads to incorrect answers.
The other setting is \textit{TextConf}, where all the KG evidence $E_{\text{KG}}$ are correct, while some textual evidence in $E_{\text{text}}$ is inconsistent with some KG evidence and leads to incorrect answers.
For convenience, we use $E^{+}_{\text{text}}$ and $E^{-}_{\text{text}}$ to denote positive textual evidence, and textual evidence leading to inconsistency (i.e., conflicting evidence), respectively. Similarly, for the KG evidence, we use $E^{+}_{\text{KG}}$ and $E^{-}_{\text{KG}}$. 
Under both settings, the LLM is expected to distinguish and exclude the conflicting evidence, and infer the correct answer.

\section{ConflictQA Benchmark}
\subsection{Benchmark Construction}
In this section, we detail the construction of the ConflictQA benchmark.
ConflictQA is built upon two widely used KG question answering datasets, i.e., WebQSP~\cite{yih2016value} and CWQ~\cite{talmor-berant-2018-web}, where each question is associated with verified golden answers grounded in the KG.
Based on how answering the question relies on different knowledge sources, we develop two settings for ConflictQA: \textit{Complementary} (COMP) and \textit{Non-complementary} (Non-COMP).
In COMP, answering a question requires jointly leveraging evidence from both textual documents and KGs, whereas in Non-COMP, evidence from either source alone is sufficient to derive an answer.
In practice, we develop these two settings by exploiting the inherent reasoning structure of the original datasets: multi-hop questions and simple single-hop questions are used to construct data of COMP and Non-COMP, respectively.

As shown in Figure~\ref{fig:bench_construct}, the construction of ConflictQA consists of two main steps: \textit{(i)} Positive Evidence Collection, where we retrieve positive evidence for each question from both the KG and Wikipedia pages, and \textit{(ii)} Conflicting Evidence Instantiation, where we employ a Conflict Generation Module to generate negative KG and textual evidence that lead to incorrect answers. 
With the question, we can directly use their positive evidence for  reasoning samples without conflict (i.e., positive reasoning scenarios), and combine positive and negative evidence for reasoning samples with conflict.

\subsubsection{Positive Evidence Collection}
The construction of ConflictQA starts with collecting positive KG and textual evidence that consistently support the correct answer for each question.
This stage establishes reliable evidence from each source as the basis for subsequent conflict construction.

\noindent \textbf{KG Evidence.}
Given a question $q$, we perform breadth-first search (BFS) on the underlying KG to retrieve triple paths that connect the entity $e_q$ mentioned in $q$ to correct answers $\mathcal{A}_q$.
These paths serve as candidate KG evidence of $q$, which needs further verification.
For example, as shown in the green box of Figure~\ref{fig:bench_construct}, the path
\textit{(Roma, filmPlace, Colonia Roma)} $\rightarrow$ \textit{(Colonia Roma, locatedIn, Mexico City)} is irrelevant to the question \textit{"What is the birthplace of the director of the movie Roma?"}, as it describes the filming location rather than the director’s birthplace.
To ensure factual correctness and relevance, three human experts independently assess whether each candidate path can validly support answering $q$.
Only paths unanimously judged as valid are retained; if all candidate paths for a question are deemed invalid, the corresponding sample is discarded.
We denote the resulting set of validated factual triples as $E_{KG}(q)$.

\noindent \textbf{Textual Evidence.}
For each question $q$, we construct textual descriptions grounded in Wikipedia that are consistent with correct answers.
Specifically, for each answer $a_i \in \mathcal{A}_q$, we extract the introductory summary paragraph from the corresponding Wikipedia page, denoted as $p(a_i)$.
Considering $p(a_i)$ may omit some answer-related facts in the KG evidence, we employ an LLM\footnote{In practice, we use GPT-4o to perform all LLM-related tasks during the construction of ConflictQA.} to check whether each triple in $E_{KG}(q)$ that involves $a_i$ is supported by $p(a_i)$.
If such triples are not supported, $p(a_i)$ is minimally rewritten by the LLM to incorporate the missing facts while preserving factual correctness.
As a result, we obtain a set of textual descriptions, each of which fully covers the information required to support its corresponding answer $a_i \in \mathcal{A}_q$.
For questions in CMOP, we additionally extract the introductory summary paragraph of the question entity $e_q$, denoted as $p(e_q)$, which is used to construct complementary cases where the textual evidence can only provide contextual information.

\noindent \textbf{Positive Reasoning Scenarios.}
With the retrieved KG triples and Wikipedia texts, we organize them for each question to construct positive reasoning scenarios.
For Non-COMP questions, which involve single-hop reasoning, all constructed evidence directly supports the correct answers.
Accordingly, all triples in $E_{KG}(q)$ form the positive KG evidence $E^+_{KG}(q)$ for a Non-COMP question $q$, while the verified texts $\{p(a_i)\}$ constitute the positive textual evidence $E^+_{\text{text}}(q)$.
In COMP, we consider two positive reasoning scenarios based on how different evidence sources cover the correct answers.
In the \textit{TriplePos} scenario, we treat $p(e_q)$ as the positive textual evidence $E^+_{\text{text}}(q)$, while the positive KG evidence $E^+_{KG}(q)$ is constructed as the subset of triples in $E_{KG}(q)$ that involve the correct answers in $\mathcal{A}_q$.
In this case, the textual evidence only provides background context, and the correct answers are primarily supported by KG triples.
In contrast, in the \textit{TextPos} scenario, we construct $E^+_{\text{text}}(q)$ using the verified answer-related texts $\{p(a_i)\}$, which explicitly cover the information required to support the correct answers. Accordingly, the positive KG evidence $E^+_{KG}(q)$ consists of the triples in $E_{KG}(q)$ that are not directly involved with the correct answers.


\subsubsection{Conflicting Evidence Instantiation}
During this stage, we first employ a Conflict Generation Module to produce negative evidence for each reasoning sample, and then instantiate conflicting reasoning scenarios: TextConf and TripleConf as introduced in Section~\ref{sec:confl}.

\noindent \textbf{Conflict Generation Module.}
This module aims to generate negative KG and textual evidence that contradicts the positive evidence.
As illustrated in the red box in Figure~\ref{fig:bench_construct}, given a question $q$, we first prompt an LLM to generate a set of semantically plausible but incorrect answers, denoted as
$\hat{\mathcal{A}}_q = \{\hat{a}_1, \hat{a}_2, \ldots, \hat{a}_n\}$,
where the number of incorrect answers $n$ matches the cardinality of the correct answer set $\mathcal{A}_q$.
These incorrect answers serve as adversarial targets for synthesizing answer-inconsistent information.
To generate negative KG triples, we modify the KG triples associated with the correct answers.
Specifically, for each $\hat{a}_i \in \hat{\mathcal{A}}_q$, we replace the correct answer entity $a_i \in \mathcal{A}_q$ in the corresponding triples with $\hat{a}_i$,
while preserving the original relational structure.
In this way, the resulting triples encode plausible but incorrect answer information.
For textual descriptions, we prompt an LLM to generate Wikipedia-style texts based on the synthesized negative KG triples.
The generated texts explicitly reflect the relational facts encoded in the negative triples, and are therefore semantically inconsistent with the corresponding positive KG evidence.

\noindent \textbf{Conflicting Reasoning Scenarios.}
For Non-COMP questions, after collecting negative evidence via the Conflict Generation Module, 
we develop TripleConf setting by combining negative KG evidence with the original positive textual evidence for each reasoning sample.
Similarly, the TextConf setting is implemented by literally pairing negative textual evidence with the corresponding positive KG evidence.
In the COMP setting, conflicting evidence is instantiated based on the predefined positive reasoning scenarios.
Specifically, under the TriplePos scenario, we additionally include negative textual evidence with original textual evidence set to form $E^-_{text}$, and keep the positive KG evidence $E^+_{KG}$ unchanged.
As a result, we construct the corresponding conflicting reasoning scenario TextConf, where answering the question still requires joint reasoning over both KG and textual evidence, whereas reasoning based solely on textual evidence leads to incorrect answers.
Similarly, under the \textit{TextPos} scenario, we form TripleConf by adding a set of negative KG triples to each question's KG evidence, without altering the positive textual evidence $E^+_{\text{text}}$.

\begin{table}[t]
\caption{Statistics of ConflictQA.}
\centering
\resizebox{0.45\textwidth}{!}{
\begin{tabular}{lccc} 
\toprule
\textbf{Conflict}  & \textbf{\#Avg Triples} & \textbf{\#Avg Tokens} & \textbf{\#Sample} \\ 
\midrule
\rowcolor{lightgray}
\multicolumn{4}{c}{\textbf{Non-Complementary}} \\
\midrule
Positive    & 1.84 & 390.41 & 802 \\
TripleConf & 1.84 & 390.41 & 802 \\
TextConf   & 1.84 & 330.82 & 802 \\ 
\midrule
\midrule
\rowcolor{lightgray}
\multicolumn{4}{c}{\textbf{Complementary}} \\
\midrule
TextPos  & 1.05 & 534.92 & 430 \\
TripleConf  & 3.57 & 534.92 & 430 \\ 
\midrule
TriplePos  & 2.54 & 50.2 & 430 \\ 
TextConf  & 2.54 & 512.93 & 430 \\

\bottomrule
\end{tabular}
}
\label{tab:bench1}
\vspace{-0.8cm}
\end{table}

\subsection{Benchmark Usage and Statistics}
ConflictQA contains both positive reasoning scenarios and their conflict-induced variants.
In evaluation, we compare model performance on positive and conflicting data to measure the performance degradation caused by cross-source conflicts.
Table~\ref{tab:bench1} summarizes the statistics of ConflictQA, including the number of samples and the distribution of KG and textual evidence across different settings.

\begin{table*}[t]
\centering
\caption{Performance (\%) of LLMs on ConflictQA under the non-complementary setting. Pos and Conf denote performance with positive and conflicting evidence, respectively. $\Delta$ indicates the resulting performance degradation. \textbf{Bold} and \underline{underline} values denote optimal and sub-optimal results, respectively. Pink cells highlight cases with substantial degradation.}
\label{tab:main1}
\resizebox{0.98\textwidth}{!}{%
\begin{tabular}{l | ccc c ccc | ccc c ccc}
\toprule

\multirow{3}{*}{\textbf{Large Language Model}} & 
\multicolumn{7}{c|}{\textbf{TripleConf}} & 
\multicolumn{7}{c}{\textbf{TextConf}} \\
\cmidrule{2-8} \cmidrule{9-15}

 & \multicolumn{3}{c}{\textbf{F1}} & & \multicolumn{3}{c|}{\textbf{EM}} & 
   \multicolumn{3}{c}{\textbf{F1}} & & \multicolumn{3}{c}{\textbf{EM}} \\

\cmidrule{2-4} \cmidrule{6-8} \cmidrule{9-11} \cmidrule{13-15}

 & \textbf{Pos} & \textbf{Conf} & $\Delta$ & & 
   \textbf{Pos} & \textbf{Conf} & $\Delta$ & 
   \textbf{Pos} & \textbf{Conf} & $\Delta$ & & 
   \textbf{Pos} & \textbf{Conf} & $\Delta$ \\
\midrule

\multicolumn{15}{c}{\textit{General LLMs}} \\
\midrule
Qwen3-8B & 
\cellcolor{lightgray}86.01 & 21.85 & \cellcolor{pink!60}\deltacell{64.16}{75} & & 
\cellcolor{lightgray}76.28 & 7.61  & \cellcolor{pink!60}\deltacell{68.67}{90} & 
\cellcolor{lightgray}86.01 & 76.24 & \deltacell{9.77}{11} & & 
\cellcolor{lightgray}76.28 & 52.44 & \deltacell{23.84}{31} \\

Llama-3.1-8B-Instruct & 
\cellcolor{lightgray}67.30 & 36.82 & \deltacell{30.48}{45} & & 
\cellcolor{lightgray}50.75 & 19.20 & \deltacell{31.55}{62} & 
\cellcolor{lightgray}67.30 & 63.39 & \deltacell{3.91}{6} & & 
\cellcolor{lightgray}50.75 & 41.15 & \deltacell{9.60}{19} \\

Llama-3.1-70B-Instruct & 
\cellcolor{lightgray}84.72 & 46.83 & \deltacell{37.89}{45} & & 
\cellcolor{lightgray}79.80 & 32.67 & \deltacell{47.13}{59} & 
\cellcolor{lightgray}84.72 & 63.10 & \deltacell{21.62}{26} & & 
\cellcolor{lightgray}79.80 & 48.96 & \deltacell{30.84}{39} \\

Open-Mistral-7B & 
\cellcolor{lightgray}81.45 & 29.70 & \cellcolor{pink!30}\deltacell{51.75}{64} & & 
\cellcolor{lightgray}65.59 & 13.47 & \cellcolor{pink!30} \deltacell{52.12}{79} & 
\cellcolor{lightgray}81.45 & 67.10 & \deltacell{14.35}{18} & & 
\cellcolor{lightgray}65.59 & 47.07 & \deltacell{18.52}{28} \\

Mistral-Large-2512 & 
\cellcolor{lightgray}84.91 & 53.65 & \deltacell{31.26}{37} & & 
\cellcolor{lightgray}78.18 & 29.30 & \deltacell{48.88}{63} & 
\cellcolor{lightgray}84.91 & 61.50 & \deltacell{23.41}{28} & & 
\cellcolor{lightgray}78.18 & 32.17 & \cellcolor{pink!60}\deltacell{46.01}{59} \\

Deepseek-V3.2 & 
\cellcolor{lightgray}\textbf{90.42} & 43.60 & \deltacell{46.82}{52} & & 
\cellcolor{lightgray}86.41 & 30.30 & \deltacell{56.11}{65} & 
\cellcolor{lightgray}\textbf{90.42} & 66.64 & \deltacell{23.78}{26} & & 
\cellcolor{lightgray}86.41 & 53.43 & \deltacell{32.98}{38} \\

GPT-3.5-Turbo-0125 & 
\cellcolor{lightgray}88.20 & 53.93 & \deltacell{34.27}{39} & & 
\cellcolor{lightgray}83.17 & \underline{36.41} & \deltacell{46.76}{56} & 
\cellcolor{lightgray}88.20 & 62.65 & \cellcolor{pink!60}\deltacell{25.55}{29} & & 
\cellcolor{lightgray}83.17 & 43.77 & \deltacell{39.40}{47} \\

GPT-4o & 
\cellcolor{lightgray}88.44 & \underline{54.39} & \deltacell{34.05}{39} & & 
\cellcolor{lightgray}85.41 & 32.67 & \deltacell{52.74}{62} & 
\cellcolor{lightgray}88.44 & \textbf{70.43} & \deltacell{18.01}{20} & & 
\cellcolor{lightgray}85.41 & 46.57 & \deltacell{38.84}{45} \\

GPT-5.1 & 
\cellcolor{lightgray}\underline{90.17} & 46.74 & \deltacell{43.43}{48} & & 
\cellcolor{lightgray}\textbf{87.28}    & 32.05 & \deltacell{55.23}{63} & 
\cellcolor{lightgray}\underline{90.17} & \underline{68.45} & \deltacell{21.72}{24} & & 
\cellcolor{lightgray}\textbf{87.28}    & 51.88 & \deltacell{35.40}{41} \\
\midrule

\multicolumn{15}{c}{\textit{Reasoning LLMs}} \\
\midrule
Qwen3-30B-A3B-Thinking & 
\cellcolor{lightgray}84.43 & 44.44 & \deltacell{39.99}{47} & & 
\cellcolor{lightgray}79.18 & 32.61 & \deltacell{46.57}{59} & 
\cellcolor{lightgray}84.43 & 60.91 & \cellcolor{pink!30}\deltacell{23.52}{28} & & 
\cellcolor{lightgray}79.18 & 48.69 & \deltacell{30.49}{39} \\

Deepseek-V3.2-Thinking & 
\cellcolor{lightgray}84.18 & \textbf{54.63} & \deltacell{29.55}{35} & & 
\cellcolor{lightgray}79.30 & 33.23 & \deltacell{46.07}{58} & 
\cellcolor{lightgray}84.18 & 62.73 & \deltacell{21.45}{25} & & 
\cellcolor{lightgray}79.30 & 40.15 & \cellcolor{pink!30}\deltacell{39.15}{49} \\

o3-mini & 
\cellcolor{lightgray}90.06 & 51.00     & \deltacell{39.06}{43} & & 
\cellcolor{lightgray}\underline{86.91} & \textbf{38.03} & \deltacell{48.88}{56} & 
\cellcolor{lightgray}90.06 & 66.66     & \deltacell{23.40}{26} & & 
\cellcolor{lightgray}\underline{86.91} & 52.56 & \deltacell{34.35}{40} \\
\bottomrule
\end{tabular}%
}
\end{table*}

\begin{table*}[t]
\centering
\caption{Performance (\%) under the complementary setting. Its settings are consistent with Table 2.}
\label{tab:main2}
\resizebox{0.98\textwidth}{!}{%
\begin{tabular}{l | ccc c ccc | ccc c ccc}
\toprule
\multirow{3}{*}{\textbf{Large Language Model}} & \multicolumn{7}{c|}{\textbf{TripleConf}} & \multicolumn{7}{c}{\textbf{TextConf}} \\
\cmidrule{2-8} \cmidrule{9-15}
 & \multicolumn{3}{c}{\textbf{F1}} & & \multicolumn{3}{c|}{\textbf{EM}} & \multicolumn{3}{c}{\textbf{F1}} & & \multicolumn{3}{c}{\textbf{EM}} \\
\cmidrule{2-4} \cmidrule{6-8} \cmidrule{9-11} \cmidrule{13-15}
 & \textbf{Pos} & \textbf{Conf} & $\Delta$ & & \textbf{Pos} & \textbf{Conf} & $\Delta$ & \textbf{Pos} & \textbf{Conf} & $\Delta$ & & \textbf{Pos} & \textbf{Conf} & $\Delta$ \\
\midrule

\multicolumn{15}{c}{\textit{General LLMs}} \\
\midrule
Qwen3-8B & 
\cellcolor{lightgray}53.26 & 26.26 & \cellcolor{pink!30}\seconddeltacell{27.00}{51} & & 
\cellcolor{lightgray}36.98 & 8.14  & \cellcolor{pink!60}\maxdeltacell{28.84}{78} & 
\cellcolor{lightgray}88.99 & \underline{68.73} & \deltacell{20.26}{23} & & 
\cellcolor{lightgray}78.63 & 46.98 & \deltacell{31.65}{40} \\

Llama-3.1-8B-Instruct & 
\cellcolor{lightgray}52.79 & 36.45 & \deltacell{16.34}{31} & & 
\cellcolor{lightgray}38.37 & 17.67 & \deltacell{20.70}{54} & 
\cellcolor{lightgray}80.39 & 42.56 & \cellcolor{pink!60}\maxdeltacell{37.83}{47} & & 
\cellcolor{lightgray}64.19 & 21.63 & \cellcolor{pink!30}\seconddeltacell{42.56}{66} \\

Llama-3.1-70B-Instruct & 
\cellcolor{lightgray}57.76 & 42.14 & \deltacell{15.62}{27} & & 
\cellcolor{lightgray}46.74 & 31.16 & \deltacell{15.58}{33} & 
\cellcolor{lightgray}83.00 & 47.46 & \cellcolor{pink!30}\seconddeltacell{35.54}{43} & & 
\cellcolor{lightgray}74.65 & 33.72 & \deltacell{40.93}{55} \\

Open-Mistral-7B & 
\cellcolor{lightgray}43.25 & 13.34 & \cellcolor{pink!60}\maxdeltacell{29.91}{69} & & 
\cellcolor{lightgray}30.70 & 6.51  & \deltacell{24.19}{79} & 
\cellcolor{lightgray}73.98 & 51.81 & \deltacell{22.17}{30} & & 
\cellcolor{lightgray}59.30 & 38.14 & \deltacell{21.16}{36} \\

Mistral-Large-2512 & 
\cellcolor{lightgray}61.01 & 51.47 & \deltacell{9.54}{16} & & 
\cellcolor{lightgray}47.91 & 29.77 & \deltacell{18.14}{38} & 
\cellcolor{lightgray}82.12 & 52.00 & \deltacell{30.12}{37} & & 
\cellcolor{lightgray}72.09 & 28.84 & \cellcolor{pink!60}\maxdeltacell{43.25}{60} \\

Deepseek-V3.2 & 
\cellcolor{lightgray}59.31 & 36.83 & \deltacell{22.48}{38} & & 
\cellcolor{lightgray}52.79 & 26.74 & \cellcolor{pink!30}\seconddeltacell{26.05}{49} & 
\cellcolor{lightgray}83.56 & 61.08 & \deltacell{22.48}{27} & & 
\cellcolor{lightgray}74.19 & \underline{51.86} & \deltacell{22.33}{30} \\

GPT-3.5-Turbo-0125 & 
\cellcolor{lightgray}63.48 & 42.53 & \deltacell{20.95}{33} & & 
\cellcolor{lightgray}50.93 & 30.70 & \deltacell{20.23}{40} & 
\cellcolor{lightgray}85.55 & 59.23 & \deltacell{26.32}{31} & & 
\cellcolor{lightgray}78.60 & 41.40 & \deltacell{37.20}{47} \\

GPT-4o & 
\cellcolor{lightgray}\underline{65.63} & \underline{50.97} & \deltacell{14.66}{22} & & 
\cellcolor{lightgray}\underline{55.35} & 33.72 & \deltacell{21.63}{39} & 
\cellcolor{lightgray}87.62 & 61.10 & \deltacell{26.52}{30} & & 
\cellcolor{lightgray}80.00 & 44.42 & \deltacell{35.58}{44} \\

GPT-5.1 & 
\cellcolor{lightgray}\textbf{66.87} & 45.85 & \deltacell{21.02}{31} & & 
\cellcolor{lightgray}\textbf{58.14} & 33.72 & \deltacell{24.42}{42} & 
\cellcolor{lightgray}\underline{89.26} & \textbf{69.18} & \deltacell{20.08}{23} & & 
\cellcolor{lightgray}\textbf{82.79} & \textbf{55.58} & \deltacell{27.21}{33} \\
\midrule

\multicolumn{15}{c}{\textit{Reasoning LLMs}} \\
\midrule
Qwen3-30B-A3B-Thinking & 
\cellcolor{lightgray}53.55 & 44.84 & \deltacell{8.71}{16} & & 
\cellcolor{lightgray}43.49 & 34.88 & \deltacell{8.61}{20} & 
\cellcolor{lightgray}84.28 & 55.35 & \deltacell{28.93}{34} & & 
\cellcolor{lightgray}76.51 & 44.42 & \deltacell{32.09}{42} \\

Deepseek-V3.2-Thinking & 
\cellcolor{lightgray}64.12 & \textbf{55.43} & \deltacell{8.69}{14} & & 
\cellcolor{lightgray}54.88 & \underline{36.98} & \deltacell{17.90}{33} & 
\cellcolor{lightgray}\textbf{89.82} & 56.34 & \deltacell{33.48}{37} & & 
\cellcolor{lightgray}\underline{81.86} & 39.30 & \deltacell{42.56}{52} \\

o3-mini & 
\cellcolor{lightgray}58.87 & 46.11 & \deltacell{12.76}{22} & & 
\cellcolor{lightgray}48.14 & \textbf{37.67} & \deltacell{10.47}{22} & 
\cellcolor{lightgray}83.70 & 60.00 & \deltacell{23.70}{28} & & 
\cellcolor{lightgray}74.65 & 48.60 & \deltacell{26.05}{35} \\
\bottomrule
\end{tabular}%
}
\end{table*}

\section{Evaluation and Analysis}
\label{sec:eva}
\subsection{LLMs for Benchmarking}
To ensure a comprehensive evaluation, we benchmark 12 popular large language models (LLMs), which can be categorized into two groups based on their intended inference characteristics.
The first group consists of general-purpose instruction-following models, referred to as \textit{General LLM} in this work.
This group includes Qwen3-8B~\cite{yang2025qwen3}, LLaMA-3.1-8B-Instruct, LLaMA-3.1-70B-Instruct~\cite{dubey2024llama}, Open-Mistral-7B~\cite{DBLP:journals/corr/abs-2310-06825}, Mistral-Large-2512~\cite{Mistral2025}, Deepseek-V3.2
~\cite{liu2025deepseek}, GPT-3.5-Turbo~\cite{openai2025gpt3}, GPT-4o~\cite{openai2025gpt4o}, and GPT-5.1~\cite{openai2025gpt51}.
The second group, denoted as \textit{Reasoning LLM}, comprises models explicitly designed to enhance multi-step reasoning capability.
They are Qwen3-30B-A3B-Thinking, Deepseek-V3.2-Thinking, and OpenAI o3-mini~\cite{openai2025gpto3}.

\vspace{-0.2cm}
\subsection{Evaluation Protocol and Metrics}
We evaluate all LLMs under a zero-shot setting, without any task-specific fine-tuning on ConflictQA.
For each question, the model is provided with the corresponding factual and textual evidence and is required to generate answers directly.
Considering the presence of conflicting information, we adopt a conflict-aware QA prompt.
Specifically, the LLM is explicitly informed that \emph{``The provided triples and texts may contain conflicting or inconsistent information''}, encouraging the model to reason cautiously.
To mitigate potential order sensitivity when presenting multiple evidence sources, we evaluate each case twice using two different evidence orders and report the averaged results when it is not specified.
Meanwhile, we particularly investigate the impact of the order of evidence, and analyse the results in Section~\ref{exp:order}.

For evaluation, we utilize the macro-F1 and Exact Match (EM) as our metrics.
Macro-F1 is computed by averaging F1 scores across questions, where the F1 score of each question is computed based on the Precision and Recall of the generated answers in comparison with the ground truth answer set, while EM is the proportion of questions for which the generated answers exactly match the ground-truth answer set.

\subsection{Performance under Knowledge Conflict}
We evaluate the performance of LLMs under both Non-COMP and COMP settings, as reported in Table~\ref{tab:main1} and Table~\ref{tab:main2}.
Our analysis mainly focuses on the following three aspects.

\subsubsection{Overall Performance}
When the evidence include knowledge conflicts, we observe a consistent and substantial performance drop across all the evaluated LLMs under both Non-COMP and COMP settings.
Notably, under the Non-COMP setting, conflicts introduced by factual evidence lead to a dramatic performance drop, with Exact Match (EM) decreasing by nearly or more than 60\% for all LLMs in TripleConf.
Under the COMP setting, the performance of LLMs also decreases pronouncedly under conflicting evidence. 
For example, under the TripleConf scenario, Qwen3-8B and Open-Mistral-7B suffer severe EM degradation, with relative drops of 78\% and 79\%, respectively.
Regrarding the TextConf setting with complementary reasoning, 9 out of the 12 evaluated LLMs exhibit relative EM degradation exceeding 40\%.
From these observations, we conclude that LLMs themselves struggle to identify and prioritize more reliable evidence for robust reasoning when facing inconsistent evidence.

\subsubsection{Comparison of Different Conflict Types}
Under both Non-COMP and COMP settings, conflicts introduced by KG evidence consistently result in more severe performance degradation than those introduced by textual evidence.
For example, despite its strong reasoning capability, GPT-5.1 suffers a performance drop of 43.43 under factual conflicts under the Non-COMP setting, which is more than twice the degradation observed under textual conflicts.
A similar trend is also observed in the COMP setting among general LLMs.
However, for reasoning LLMs, the relative performance drop caused by negative KG evidence is conversely weaker than that caused by negative textual evidence.

These results suggest that LLMs tend to rely more heavily on concise and explicit triple-based information during reasoning.
Consequently, corrupting factual evidence leads to pronounced failures, whereas conflicting textual evidence generally has a weaker impact on the final prediction.

\subsubsection{Comparison of Different LLMs}
\label{exp:llm}
There is considerable variation among different LLMs in their robustness to conflicting evidence.
Overall, reasoning LLMs achieve better performance than general instruction-following LLMs under most conflict scenarios, demonstrating improved faithfulness to cross-source inconsistencies.
However, this advantage is not uniform across all conditions.
In particular, under complementary scenarios with textual conflicts, we observe cases where Deepseek-V3.2-Thinking degrades more than its non-thinking counterpart Deepseek-V3.2. 
This is likely because reasoning LLMs rely more heavily on textual information during reasoning, making them more susceptible to misleading signals in conflicting textual evidence.
Moreover, in general LLMs, we find that LLMs with fewer parameters, such as Qwen3-8B and Open-Mistral-7B, exhibit substantially larger performance degradation across multiple conflict settings, indicating limited robustness to inconsistent information.
Within the same model family (e.g., the LLaMA-3.1 series), larger model sizes generally lead to improved reasoning results in the presence of conflicting evidence.
Meanwhile, it is worth noting that although LLaMA-3.1-8B-Instruct exhibits the smallest performance drop under the sufficient setting with textual conflicts, this behavior is largely attributable to its relatively weak performance on positive instances, rather than the robust ability to handle conflicting evidence.

\begin{table}[t]
\centering
\caption{Performance gap caused by evidence ordering in the non-complementary setting. Light-blue cells indicate negative gaps.}
\label{tab:order}
\resizebox{0.98\linewidth}{!}{%
\begin{tabular}{l|cccc}
\toprule
\multirow{2}{*}{\textbf{Large Language Model}} & \multicolumn{2}{c}{\textbf{TripleConf}} & \multicolumn{2}{c}{\textbf{TextConf}} \\
 & \small $\Delta$ F1 & \small $\Delta$ EM & \small $\Delta$ F1 & \small $\Delta$ EM \\
\midrule

\midrule
Qwen3-8B                & 19.74 & 8.73  & \cellcolor{lightblue}-0.30 & \cellcolor{lightblue}-0.87 \\ 
Llama-3.1-8B-Instruct    & \cellcolor{lightblue}-4.72 & 4.74  & \cellcolor{lightblue}-0.33 & \cellcolor{lightblue}-9.48 \\
Llama-3.1-70B-Instruct & \cellcolor{lightblue}-1.41 & \cellcolor{lightblue}-0.50 & \cellcolor{lightblue}-4.42 & \cellcolor{lightblue}-6.86 \\
Open-Mistral-7B         & 9.79  & 8.48  & 7.87 & 8.10 \\
Mistral-Large-2512      & 1.93  & 6.98  & 5.12 & \cellcolor{lightblue}-1.74 \\ 
Deepseek-V3.2           & \cellcolor{lightblue}-7.71 & \cellcolor{lightblue}-3.74 & \cellcolor{lightblue}-3.32 & \cellcolor{lightblue}-7.60 \\
GPT-3.5-Turbo-0125      & 0.63  & 4.98  & \cellcolor{lightblue}-3.30 & 6.49  \\
GPT-4o                  & 4.14  & 10.48 & 0.74  & 1.62  \\
GPT-5.1                 & \cellcolor{lightblue}-4.82 & \cellcolor{lightblue}-0.75 & 0.49 & 0.25 \\ 
\midrule

\midrule
Qwen3-30B-A3B-Thinking  & \cellcolor{lightblue}-5.08 & \cellcolor{lightblue}-3.61 & \cellcolor{lightblue}-0.15 & \cellcolor{lightblue}-1.12 \\ 

Deepseek-V3.2-Thinking       & \cellcolor{lightblue}-4.20 & \cellcolor{lightblue}-2.62 & \cellcolor{lightblue}-1.53 & \cellcolor{lightblue}-4.48 \\ 
o3-mini                 & \cellcolor{lightblue}-0.65 & 0.24 & 0.20  & 0.87  \\ 
\bottomrule
\end{tabular}%
}
\end{table}

\begin{figure*}[t]
  \vspace{-0.2cm}
  \centering
  \includegraphics[width=0.95\linewidth]{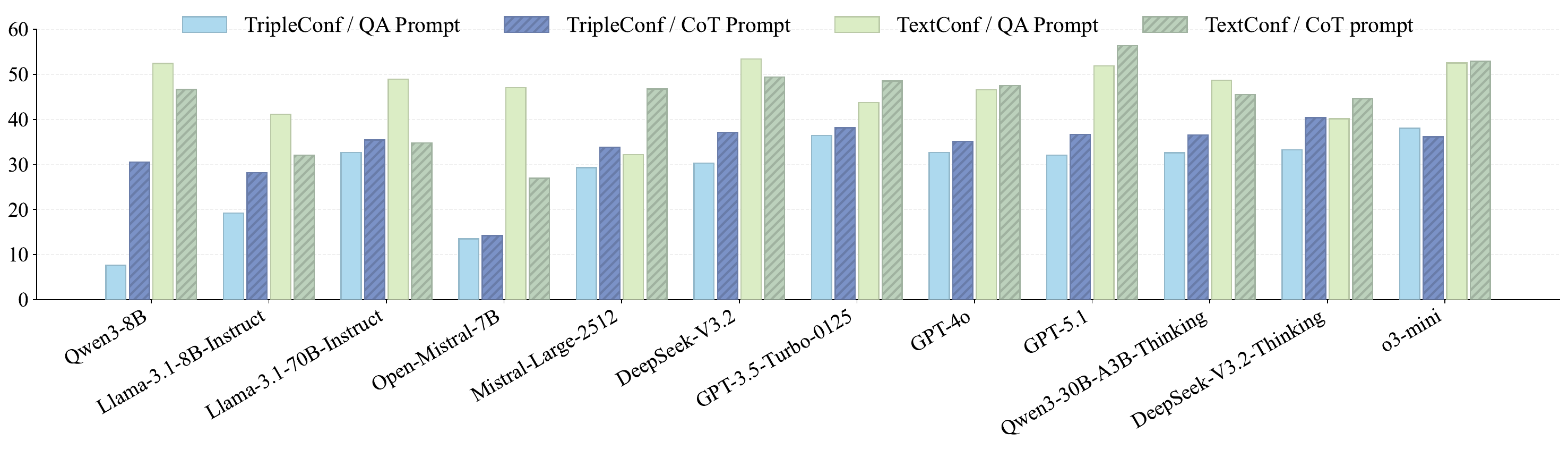}
  \caption{Exact Match results (\%) of LLMs utilizing different prompts under non-complementary reasoning scenarios.}
  \Description{}
  \label{fig:cot_1}
\end{figure*}

\begin{figure*}[t]
  \centering
  \includegraphics[width=0.95\linewidth]{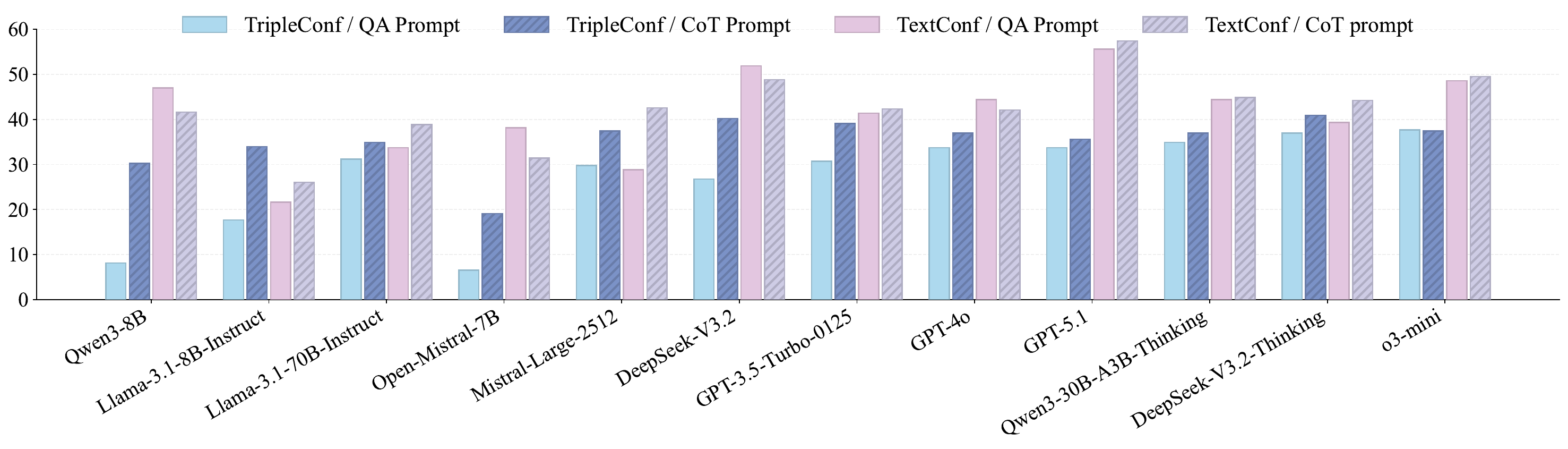}
  \caption{Exact Match results (\%) of LLMs utilizing different prompts under complementary reasoning scenarios.}
  \Description{}
  \label{fig:cot_2}
\end{figure*}

\subsection{Effect of Evidence Ordering}\label{exp:order}
In this section, we study how evidence ordering affects LLM performance with the direct conflict-aware prompt under the Non-COMP setting.
Specifically, we record each LLM's results when the conflicting evidence is present before and after the correct evidence, and analyze the performance gap.

From the results shown in Table~\ref{tab:order}, we observe that the performance gap varies considerably across models and conflict types.
Overall, changing the evidence order does not consistently enable faithful reasoning across all LLMs.
For LLMs that are particularly vulnerable to conflicting evidence, such as Qwen3-8B and Open-Mistral-7B (discussed in Section~\ref{exp:llm}), placing correct evidence before conflicting evidence can lead to notable performance improvements.
When comparing the two categories of LLMs, reasoning LLMs generally exhibit smaller performance gaps, indicating that they are less affected by evidence ordering, whereas general LLMs are more sensitive to the order of evidence.
Among all models, o3-mini demonstrates the greatest stability, exhibiting consistently smaller gaps across different conflict types.

\subsection{Investigation on Prompt Strategy}
\label{exp:prompt}
We also evaluate whether the CoT prompting strategy can lead to better performance under conflicting evidence, in comparison with the direct QA prompt.
Through the results in Figures~\ref{fig:cot_1} and~\ref{fig:cot_2}, we have the following observations.
(i) Under both COMP and Non-COMP, CoT improves EM scores for 11 out of 12 LLMs under the TripleConf setting, with the exception of o3-mini, which exhibits a slight performance degradation.
(ii) When negative textual evidence is provided, the effectiveness of CoT varies across models.
LLMs with weaker reasoning capabilities, such as Qwen3-8B and Open-Mistral-7B, tend to be more easily influenced by negative textual evidence and consequently make more errors.
In contrast, stronger models, including GPT-5.1 and other reasoning LLMs, continue to benefit from CoT.
(iii) Reasoning LLMs exhibit smaller performance changes than general LLMs when switching from direct QA to CoT prompting.
This suggests that, even without explicit reasoning instructions, reasoning LLMs may already perform implicit step-by-step reasoning.

To better understand model behavior under CoT prompting, we further analyze the reasoning details exhibited in TextConf settings.
We observe that LLMs frequently justify their decisions by explicitly referring to the \textit{“detailed context provided in the text”}, which leads them to prioritize textual descriptions that appear more informative or comprehensive when resolving cross-source conflicts.
Even when such textual evidence is misleading, models may implicitly treat it as more reliable than concise factual triples.
This finding helps explain why CoT prompting does not consistently improve LLMs' performance under textual conflicts.

\section{Two-stage Explanation-based Thinking (XoT)}
As illustrated in the previous analysis, LLMs often exhibit implicit preferences when reasoning over conflicting evidence, which may bias models toward relying on particular types of evidence without distinguishing the conflict.
To mitigate this issue, we introduce XoT, a simple and model-agnostic two-stage explanation-based thinking framework.
In this section, we first describe the details of XoT, and then present its experimental results and case study on ConflictQA.

\begin{table*}[t]

\centering
\caption{Performance (\%) of XoT on the ConflictQA benchmark. \textbf{Bold} and \underline{underline} values denote optimal and sub-optimal results, respectively. Cells in pink indicate performance improvements of XoT over the conflict-aware QA prompt.}
\label{tab:XoT}
\resizebox{0.98\linewidth}{!}{%
\begin{tabular}{l c | cc c cc | cc c cc}
\toprule
\multirow{3}{*}{\textbf{Large Language Models}} & \multirow{3}{*}{} & \multicolumn{5}{c|}{\textbf{Non-COMP}} & \multicolumn{5}{c}{\textbf{COMP}} \\
\cmidrule{3-12}
 & & \multicolumn{2}{c}{\textbf{TripleConf}} & & \multicolumn{2}{c|}{\textbf{TextConf}} & \multicolumn{2}{c}{\textbf{TripleConf}} & & \multicolumn{2}{c}{\textbf{TextConf}} \\
\cmidrule{3-4} \cmidrule{6-7} \cmidrule{8-9} \cmidrule{11-12}
 & & \textbf{F1} & \textbf{EM} & & \textbf{F1} & \textbf{EM} & \textbf{F1} & \textbf{EM} & & \textbf{F1} & \textbf{EM} \\
\midrule

\multirow{2}{*}{Qwen3-8B} & XoT & 46.07 & 13.34 & & 55.35 & 21.57 & 45.10 & 15.58 & & 61.06 & 23.72 \\
 & $\Delta$ & \cellcolor{pink!60}24.22$_{111\%}$ $\uparrow$ & \cellcolor{pink!30}5.73$_{75\%}$ $\uparrow$ & & 20.89$_{27\%}$ $\downarrow$ & 30.87$_{59\%}$ $\downarrow$ & \cellcolor{pink!30} 18.87$_{72\%}$ $\uparrow$ & \cellcolor{pink!60}7.44$_{91\%}$ $\uparrow$ & & 7.67$_{11\%}$ $\downarrow$ & 23.26$_{50\%}$ $\downarrow$ \\
\midrule

\multirow{2}{*}{Open-Mistral-7B} & XoT & 36.88 & 9.35 & & 48.01 & 18.45 & 37.80 & 13.02 & & 51.02 & 23.26 \\
 & $\Delta$ & \cellcolor{pink!30}7.19$_{24\%}$ $\uparrow$ & 4.12$_{31\%}$ $\downarrow$ & & 19.09$_{28\%}$ $\downarrow$ & 28.62$_{61\%}$ $\downarrow$ & \cellcolor{pink!60}24.46$_{183\%}$ $\uparrow$ & \cellcolor{pink!30}6.51$_{100\%}$ $\uparrow$ & & 0.79$_{2\%}$ $\downarrow$ & 14.88$_{39\%}$ $\downarrow$ \\
\midrule
\midrule

\multirow{2}{*}{GPT-3.5-Turbo-0125} & XoT & \underline{59.35} & \underline{39.78} & & \underline{63.51} & \underline{40.40} & 54.59 & 39.77 & & \underline{64.14} & 41.86 \\
 & $\Delta$ & \cellcolor{pink!30}5.43$_{10\%}$ $\uparrow$ & \cellcolor{pink!30}\cellcolor{pink!30}3.37$_{9\%}$ $\uparrow$ & & \cellcolor{pink!30}0.86$_{1\%}$ $\uparrow$ & 3.37$_{8\%}$ $\downarrow$ & \cellcolor{pink!30}12.06$_{28\%}$ $\uparrow$ & \cellcolor{pink!30}9.07$_{30\%}$ $\uparrow$ & & \cellcolor{pink!30}4.91$_{8\%}$ $\uparrow$ & \cellcolor{pink!30}0.46$_{1\%}$ $\uparrow$ \\
\midrule

\multirow{2}{*}{GPT-4o} & XoT & \textbf{63.31} & \textbf{43.14} & & \textbf{63.68} & \textbf{41.90} & \textbf{60.99} & \textbf{42.79} & & \textbf{66.97} & \textbf{46.51} \\
 & $\Delta$ & \cellcolor{pink!30}8.92$_{16\%}$ $\uparrow$ & \cellcolor{pink!60}10.47$_{32\%}$ $\uparrow$ & & 6.75$_{10\%}$ $\downarrow$ & 4.67$_{10\%}$ $\downarrow$ & \cellcolor{pink!30}10.02$_{20\%}$ $\uparrow$ & \cellcolor{pink!30}9.07$_{27\%}$ $\uparrow$ & & \cellcolor{pink!60}5.87$_{10\%}$ $\uparrow$ & \cellcolor{pink!30}2.09$_{5\%}$ $\uparrow$ \\
\midrule
\midrule

\multirow{2}{*}{Deepseek-V3.2-Thinking} & XoT & 59.32 & 36.53 & & 59.30 & 36.53 & \underline{57.60} & \underline{42.56} & & 61.81 & 43.72 \\
 & $\Delta$ & \cellcolor{pink!30}4.69$_{9\%}$ $\uparrow$ & \cellcolor{pink!30}3.30$_{10\%}$ $\uparrow$ & & 3.43$_{6\%}$ $\downarrow$ & 3.62$_{9\%}$ $\downarrow$ & \cellcolor{pink!30}2.17$_{4\%}$ $\uparrow$ & \cellcolor{pink!30}5.58$_{15\%}$ $\uparrow$ & & \cellcolor{pink!30}5.47$_{10\%}$ $\uparrow$ & \cellcolor{pink!60}4.42$_{11\%}$ $\uparrow$ \\
\midrule

\multirow{2}{*}{o3-mini} & XoT & 57.38 & 39.78 & & 57.82 & 38.78 & 53.35 & 40.23 & & 62.19 & \underline{46.28} \\
 & $\Delta$ & \cellcolor{pink!30}6.39$_{13\%}$ $\uparrow$ & \cellcolor{pink!30}1.75$_{5\%}$ $\uparrow$ & & 8.84$_{13\%}$ $\downarrow$ & 13.77$_{26\%}$ $\downarrow$ & \cellcolor{pink!30} 7.24$_{16\%}$ $\uparrow$ & \cellcolor{pink!30}2.56$_{7\%}$ $\uparrow$ & & \cellcolor{pink!30}2.19$_{4\%}$ $\uparrow$ & 2.32$_{5\%}$ $\downarrow$ \\
\bottomrule
\end{tabular}
}
\end{table*}

\vspace{-0.2cm}
\subsection{Method Details}\label{sec:xotdetails}
Rather than straightforward LLM prompting, XoT reorganizes the roles of LLMs into a two-stage reasoning process, separating answer exploration from final decision making.

Specifically, given a question $q$ with heterogeneous evidence $(E_{\mathrm{KG}}, E_{\mathrm{text}})$, XoT first prompts the model to enumerate a set of candidate answers $\mathcal{C}_q = \{c_1, \ldots, c_K\}$, without performing correctness judgment or pruning candidates based on potential contradictions among the evidence.
For each candidate $c_i \in \mathcal{C}_q$, the LLM is further required to generate a short, answer-conditioned explanation $exp_i$ grounded in the provided evidence, while avoiding explicit references to evidence sources.
By explicitly expanding the answer space and associating each candidate with an independent explanation, XoT leaves the conflict judgment to the subsequent stage.
After obtaining the pairs of candidate answer and explanation $\{(c_i, exp_i)\}_{i=1}^{K}$ for a question $q$, 
XoT prompts the model to think over the potential conflicts among candidate explanations and infer the most appropriate final answers.

XoT serves as a simple yet effective baseline for mitigating implicit preferences in LLM reasoning under conflicting evidence.
It is worth noting that XoT can be naturally extended with more sophisticated modules, such as iterative refinement or rethinking mechanisms.
We leave such extensions for future investigation.

\subsection{Results on ConflictQA}
We test XoT on the evaluated LLMs on the ConflictQA benchmark.
Following the analysis in Section~\ref{sec:eva}, we report detailed results on six representative models with different levels of robustness under conflicting evidence.

As shown in Table~\ref{tab:XoT}, XoT improves performance in most cases, with particularly stable gains under TripleConf across both Non-COMP and COMP settings.
The improvements are especially pronounced for models that are more sensitive to conflicts.
For instance, in non-complementary settings, Qwen3-8B achieves a relative improvement of 111\% in F1,
while under complementary settings, Open-Mistral-7B obtains relative gains of 183\% in F1 and 100\% in EM.
In contrast, the effects of XoT under TextConf are more model-dependent.
Models with limited capacity to assess the reliability of textual evidence may still be misled by incorrect descriptions, leading to smaller gains or even performance degradation in some cases.
Meanwhile, stronger models continue to benefit from XoT under the COMP and TextConf settings.
This indicates that, without introducing additional supervision or external evidence, and solely relying on the LLMs’ inherent knowledge and reasoning ability, misleading textual descriptions can still dominate the reasoning. As a result, models struggle to correctly assess conflicting explanations or to select the correct answer at the final judgment stage.

For general LLMs with stronger ability, we observe that GPT-4o equipped with XoT achieves the best overall performance across all evaluated settings.
Notably, GPT-4o consistently outperforms other models, and in several settings attains results that are comparable or better than those reasoning LLMs.
GPT-3.5-Turbo-0125 also shows clear improvements with XoT, achieving the second-best performance in most cases.
These observations suggest that XoT aligns particularly well with strong general LLMs.
Despite their relatively higher robustness to conflicting evidence, reasoning models (e.g., Deepseek-V3.2-Thinking and o3-mini) also benefit from XoT, although the improvements are generally smaller in magnitude.
This is consistent with the fact that such models often exhibit structured reasoning even under direct QA prompting, leaving less space for additional gains from re-organizing the inference procedure.

Overall, the experimental results demonstrate that XoT can well serve as a model-agnostic baseline for mitigating conflict-induced performance degradation on ConflictQA, while also highlighting that resolving textual conflicts remains challenging.

\subsection{Case Study}
\begin{figure}[t]
  \centering
  \includegraphics[width=0.98\linewidth]{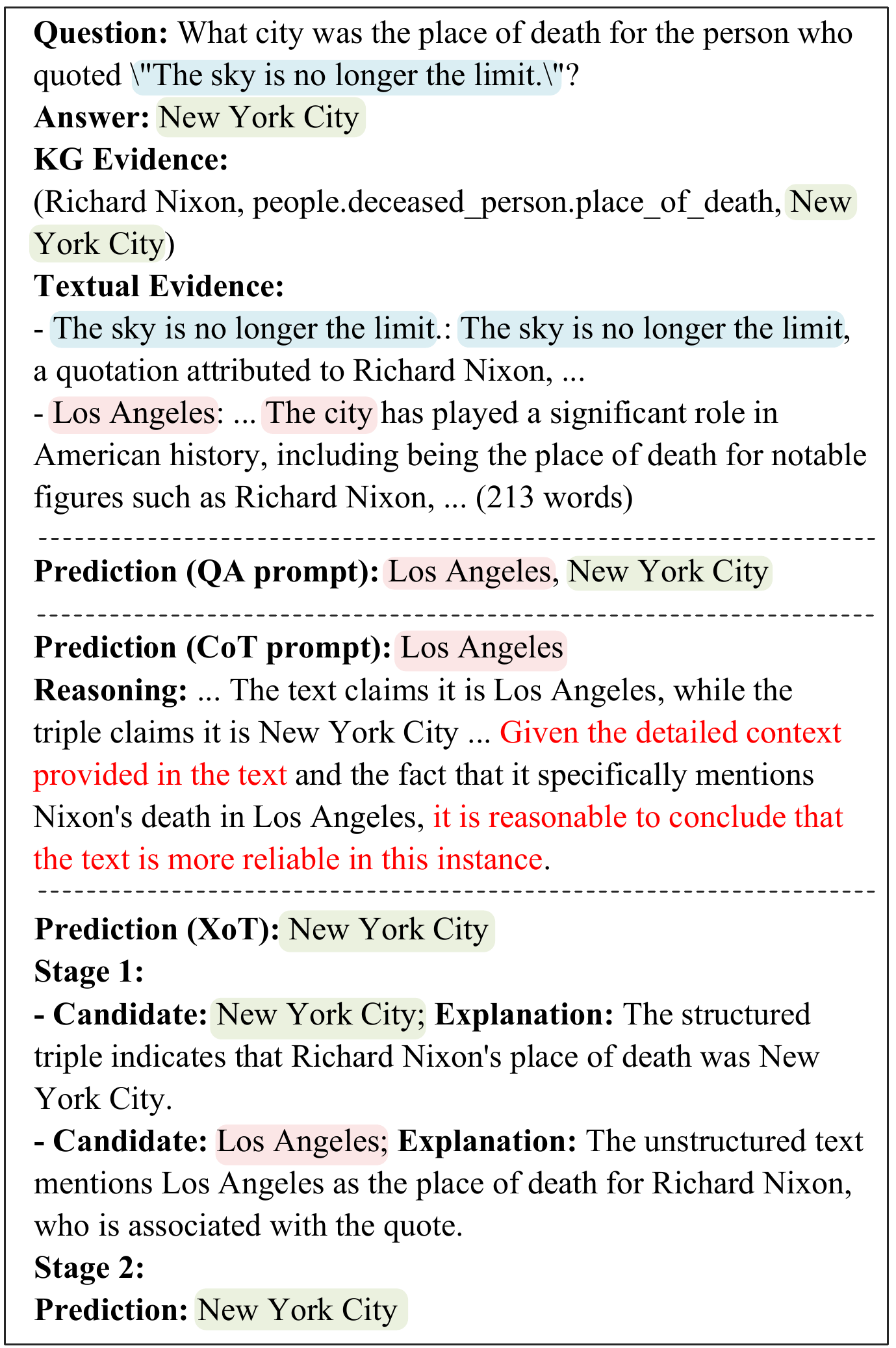}
  \caption{An example illustrating the predictions of GPT-4o under different reasoning strategies on ConflictQA in the COMP setting with misleading textual evidence.
           Entities highlighted in green and pink denote the golden answer and the misleading answer, respectively.}
  \Description{}
  \label{fig:case}
  \vspace{-0.6cm}
\end{figure}

We present a case to provide an explicit illustration of the predictions produced by different reasoning strategies under conflicting evidence.
As shown in Figure~\ref{fig:case}, under the direct QA prompt, GPT-4o treats all provided evidence as plausible and consequently outputs both correct and incorrect answers.
When prompted with CoT, GPT-4o exhibits greater trust in the detailed and narrative-rich text, explicitly judging it to be more reliable than the positive KG evidence, and consequently predicts only the misleading answer.
This observation is further consistent with the discussion in Section~\ref{exp:prompt}.
Regarding XoT, reasoning benefits from separating evidence comparison from answer generation, resulting in a correct prediction.
Overall, this case illustrates the effectiveness of XoT in resolving cross-source conflicts during LLM reasoning.

\section{Related Work}

\noindent\textbf{LLM Reasoning with RAG}.
Recent studies have explored enhancing LLM reasoning by RAG for incorporating different knowledge sources including unstructured text and (semi-)structured data like knowledge graphs and tables~\cite{wu2024stark, liu-etal-2025-ska, christmann2024rag, ma2025unifying, jeong2025database, ma2025thinkongraph}.
Among these settings, combining document with KG has attracted particular attention, as it naturally bridges unstructured and structured information and can augment the reasoning via the structured knowledge~\cite{wu2024stark}.
Existing works~\cite{ma2025thinkongraph, lee2025hybgrag, li2024chainofknowledge, xia2025knowledge} predominantly focus on designing retrieval strategies for effective evidence and developing prompting and training methods for accurate LLM reasoning.
They typically assume that the retrieved evidence is reliable and mutually consistent, ignoring potential inconsistencies introduced by the retrieval procedure, or quality issues of the sources like expiration.
Some recent studies~\cite{nazary2025poison, zou2025poisonedrag, zhao2025exploring} have revealed the vulnerability of such RAG and LLM-based systems when external knowledge sources are of low quality or have been adversarially manipulated.
In such cases, corrupted or misleading evidence can amplify conflicts across retrieved contexts and significantly distort the model’s reasoning process~\cite{wang2025astute}.
Therefore, investigating conflicting evidence becomes critical and urgent for faithful LLM reasoning with RAG.

\noindent\textbf{Knowledge Conflict in LLM Reasoning}.
The risk of conflicting or inconsistent information brought by retrieved evidence has recently garnered significant attention among LLM and RAG researchers~\cite{chen-etal-2022-rich, xie2023adaptive, xu-etal-2024-knowledge-conflicts}.
Xie et al.~\cite{xie2023adaptive} analyze conflicts between parametric knowledge and external evidence by introducing LLM-generated coherent counter-memory that explicitly contradicts elicited parametric answers.
They show that when both supportive and contradictory evidence to the LLMs’ internal knowledge are provided, LLMs can become highly receptive to externally supplied information.
Jin et al.~\cite{jin-etal-2024-tug} study knowledge conflicts in RAG by inducing LLMs’ parametric memory via closed-book QA and distilling LLM-generated counterfactual answers together with coherent conflicting evidence from existing QA datasets.
Their experiments reveal that stronger LLMs often persist in relying on incorrect internal memory even when correct external evidence is available, and that models generally prefer evidence consistent with their prior internal beliefs.
ConflictBank~\cite{su2024textttconflictbank} is a benchmark designed to evaluate conflicts between LLMs’ inherent knowledge and retrieved contextual knowledge, categorizing such conflicts into three fine-grained types: misinformation, temporal, and semantic conflicts.
These works focus on conflicts between internal LLM parameters and external evidence, but ignore the conflict between external evidence.
FaithEval~\cite{ICLR2025_48404cd9} attempts to benchmark conflicts among retrieved contexts, but only synthesize conflicts within one single source. 
%
To the best of our knowledge, there is a short of benchmarks that models different kinds of conflicts among multiple external sources.
Our benchmark ConflictQA, equipped with different conflict settings, bridges this gap and leads to systematic evaluation and results on 12 different LLMs and 3 different prompting strategies (including XoT proposed by ourselves).

\section{Conclusion and Future Work}
This work is among the first to investigate LLM reasoning under conflicting evidence retrieved from multiple knowledge sources.
We introduced ConflictQA, a novel benchmark that explicitly instantiates conflicts between textual and KG evidence under four different settings.
Through comprehensive evaluations on 12 representative LLMs with both direct and CoT prompts, we observed that resolving cross-source conflicts remains a significant challenge, as LLMs often fail to distinguish conflicting evidence, and LLMs become much more sensitive to prompt designs and evidence presentation facing evidence conflicts.
For more robust LLM reasoning with conflicting evidence, we further proposed XoT, a model-agnostic prompting strategy that asks LLMs to think in two stages with candidate ansers and explanations, and demonstrated its effectiveness on ConflictQA.
In the future, we plan to extend ConflictQA to cover more external sources, and improve XoT with other strategies like iterative refinement or rethinking as discussed in Section \ref{sec:xotdetails}.


\bibliographystyle{ACM-Reference-Format}
\bibliography{sample-base}


\end{document}